\documentclass[10pt,journal,compsoc]{IEEEtran}

\usepackage{eccvabbrv}
\usepackage{cite}
\usepackage{amsmath}
\usepackage{amsfonts}
\usepackage{amssymb}
\usepackage{algorithm}
\usepackage{algorithmic}
\usepackage{array}
\usepackage{textcomp}
\usepackage{stfloats}
\usepackage{url}
\usepackage{verbatim}
\usepackage{graphicx}
\usepackage{booktabs}
\usepackage{multirow}
\usepackage{makecell}
\usepackage[table]{xcolor}
\usepackage{colortbl}
\usepackage{adjustbox}
\usepackage{wrapfig}
\usepackage{tcolorbox}
\usepackage[caption=false,font=normalsize,labelfont=sf,textfont=sf]{subfig}
\usepackage[accsupp]{axessibility}
\usepackage{hyperref}

\newcommand{\best}[1]{\textcolor{red}{\textbf{#1}}}
\newtheorem{theorem}{Theorem}
\hypersetup{
  colorlinks=true,
  pdfborder={0 0 0},
  citecolor=blue,
  linkcolor=blue,
  urlcolor=blue
}

\begin{document}
\sloppy
\title{SAB-LVLM: Significance-Aware Binarization for Large Vision-Language Models} 

% TODO FINAL: Add corresponding author information if needed.
\author{Qi Lyu\textsuperscript{1,2,3$\dagger$}, Jiahua Dong\textsuperscript{4$\dagger$},
Baichen Liu\textsuperscript{1,2}, Xudong Wang\textsuperscript{1,2,3},
Mingfei Han\textsuperscript{4}, Yulun Zhang\textsuperscript{5},\\
Fahad Shahbaz Khan\textsuperscript{4}, Salman Khan\textsuperscript{4},
Lianqing Liu\textsuperscript{1,2}, and Zhi Han\textsuperscript{1,2} \\
% \vspace{-3mm}
$\;$ \\
$^{1}$State Key Laboratory of Robotics and Intelligent Systems \\
$^{2}$Shenyang Institute of Automation, Chinese Academy of Sciences \\
$^{3}$University of Chinese Academy of Sciences \\
$^{4}$Mohamed bin Zayed University of Artificial Intelligence \\
$^{5}$Shanghai Jiao Tong University \\
\vspace{-8mm}

\thanks{\textsuperscript{$\dagger$}Qi Lyu and Jiahua Dong contributed equally to this work.}%
}

\markboth{IEEE Computer Society Journal,~Vol.~XX, No.~X, 2026}%
{Qi Lyu \MakeLowercase{\textit{et al.}}: SAB-LVLM}

\IEEEaftertitletext{\vspace{-8mm}}
\maketitle

\begin{abstract}
Large Vision-Language Models (LVLMs) have achieved remarkable progress in multimodal understanding, yet their enormous parameter scale and cross-modal computation incur substantial memory and latency overhead, severely limiting real-world deployment on resource-constrained devices. Binarization offers an attractive solution by drastically reducing storage and computational costs. However, existing binarization methods neglect the varying importance of weights across different layers and modalities. This causes parameters irrelevant to downstream tasks to be unnecessarily retained, whereas modality-critical weights may not be adequately optimized, resulting in significant performance degradation. To address these challenges, we develop a novel \underline{S}ignificance-\underline{A}ware \underline{B}inarization for \underline{L}arge \underline{V}ision-\underline{L}anguage \underline{M}odels (SAB-LVLM). 
Specifically, after constructing Hessian matrices for textual and visual inputs, we propose a spatial significance map to distinguish full-precision weights activated under a single modality from those activated across modalities. We then devise a modality-guided integration strategy to obtain the significance-aware binarization map, which measures weight significance across layers and modalities. Subsequently, this binarization map is incorporated into the binarization objective as an error reweighting term, and binarization fitting is performed through an alternating significance-weighted update scheme.
Extensive experiments illustrate the superiority of our SAB-LVLM over existing binary PTQ methods under an approximately 1-bit compression constraint. Our code is accessible at \url{https://github.com/LyuQi127/SAB_LVLM}.
\end{abstract}


\vspace{-3mm}
\section{Introduction}\label{sec:intro}
In recent years, large language models (LLMs) built upon the transformer architecture have achieved state-of-the-art results on a broad spectrum of natural language processing tasks~\cite{deepseekai2025deepseekv3technicalreport}. Their strong empirical performance primarily stems from their enormous parameter counts, which frequently reach tens of billions. For instance, the open pretrained transformer (OPT) family~\cite{Zhang2022OPTOP} provides variants scaling up to 66 billion parameters, while the LLaMA series~\cite{touvron2023llamaopenefficientfoundation} includes even larger versions such as LLaMA3-70B. Despite their effectiveness, these models impose an immense memory footprint and computational burden. For example, running a 70B model in full precision requires over 130 GB of memory, which poses serious challenges for deployment on resource-constrained mobile devices and restricts the widespread adoption of LLMs.

To address these limitations, numerous compression techniques~\cite{frantar-gptq,Dong2024STBLLMBT,li2025mbq} have been developed for LLMs, such as weight quantization~\cite{SmoothQuant,Kim2023SqueezeLLMDQ}, network pruning~\cite{sun2024wanda,gao2024displlm}, knowledge distillation~\cite{liu2024ddk,gu2024minillm}, and low-rank factorization~\cite{saha2024compressing,zhang2024loraprune}. Among them, binarization is particularly notable since it reduces each parameter to approximately one bit, achieving an extremely high level of storage efficiency~\cite{chen2025hbllm}. The post-training quantization (PTQ) framework enables the conversion of full-precision models into low-bit counterparts without incurring the substantial cost of retraining. As a PTQ approach, binarization enables approximately 1-bit weight quantization. Recent approaches~\cite{yuan2024pbllm,huang2024billm,li2025arbllm} within this framework, including PB-LLM~\cite{yuan2024pbllm} and BiLLM~\cite{huang2024billm}, aim to mitigate performance degradation by identifying parameters that are most influential to model accuracy and applying refined optimization strategies or selective higher-precision representations, thus striking a more favorable trade-off between compression efficiency and task performance. Additionally, ARB-LLM~\cite{li2025arbllm} introduces an alternating refinement strategy for binarization that iteratively adjusts the binary parameters to substantially decrease the quantization error.

\begin{figure*}[t]
    \centering
    \subfloat[Comparison between ours method and other methods.]{
        \includegraphics[height=4.8cm]{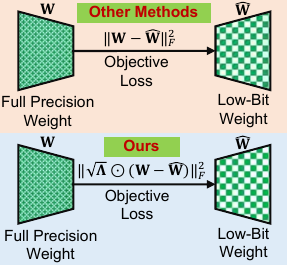}
    }\hfill
    \subfloat[Visualization results of $\mathbf{\Gamma}$ and $\mathbf{\Lambda}$.]{
        \includegraphics[height=4.8cm]{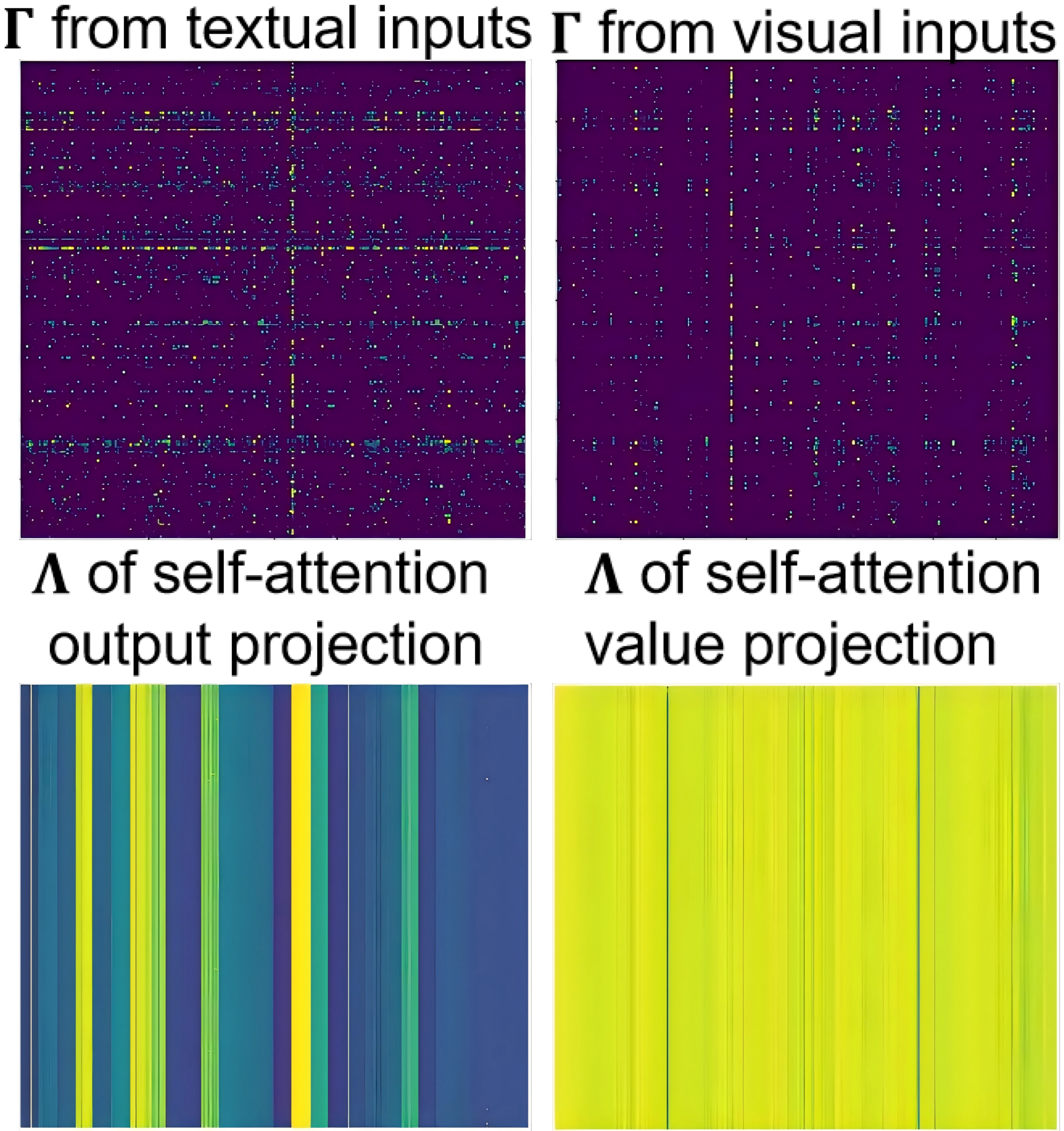}
    }\hfill
    \subfloat[Performance comparison on the MMStar Benchmark.]{
        \includegraphics[height=4.8cm]{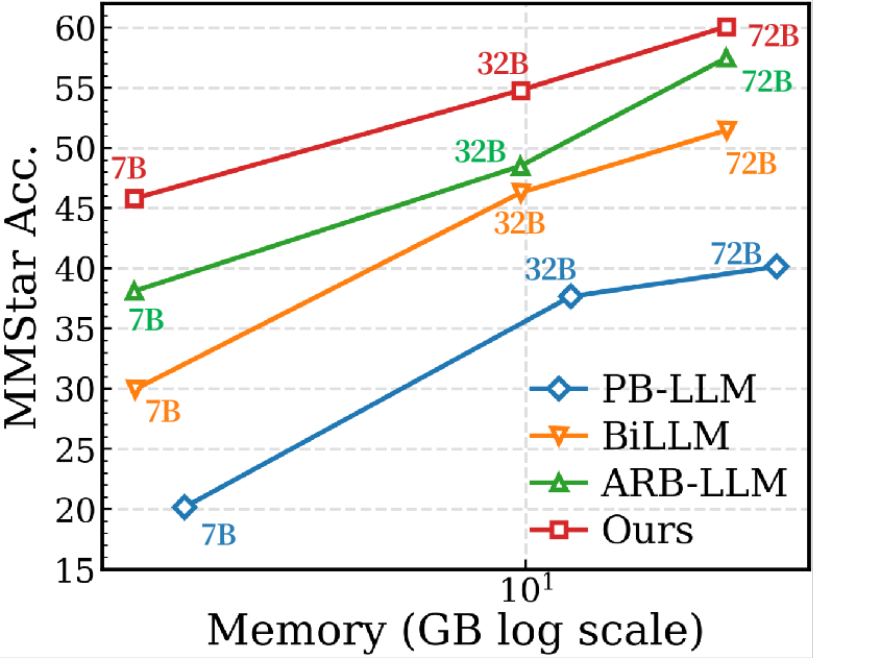}
    }
    \vspace{-2mm}
    \caption{(a): Comparison between the proposed SAB-LVLM and the other methods.
    (b): The top is the visualization results of spatial significance map $\mathbf{\Gamma}$ from textual inputs and visual inputs at the $0$-th self-attention query projection. The bottom is the visualization results of $0$-th self-attention output projection and self-attention value projection.
    (c): Comparison of performance across the Qwen2.5-VL family on the MMStar benchmark.
    }
    \vspace{-5mm}
\label{fig:motivation}
\end{figure*}

Despite these advances, it remains challenging to directly extend existing PTQ methods to large vision-language models (LVLMs). Unlike LLMs, LVLMs integrate visual encoders with large language backbones, requiring precise cross-modal alignment between visual and textual representations. Under such settings, model parameters from different layers and modalities contribute unevenly to downstream multimodal tasks. However, most PTQ approaches focus solely on minimizing the global weight quantization error after post-training, overlooking the heterogeneous importance of weights across layers and modalities, as shown in Fig. \ref{fig:motivation}(a)(b). As a result, task-irrelevant encoding parameters may be mistakenly preserved, while modality-critical weights may be insufficiently optimized, leading to severe performance degradation under weight binarization.

To address the aforementioned challenges, we propose a novel \underline{S}ignificance-\underline{A}ware \underline{B}inarization for \underline{L}arge \underline{V}ision-\underline{L}anguage \underline{M}odels (\textbf{SAB-LVLM}), which represents an early attempt at achieving binarization for LVLMs. \textbf{Firstly}, we separately feed textual and visual calibration data into LVLMs. Based on the corresponding hidden states, we construct Hessian matrices to evaluate the significance of each full-precision weight. \textbf{Secondly}, we employ a spatial significance map to distinguish weights activated under a single modality from those activated by inputs from different modalities. Furthermore, we perform modality-guided significance integration using a modality integration score, yielding the final significance-aware binarization map. Unlike existing PTQ approaches~\cite{bai2025skim,wang2026sliderquant}, as presented in Fig.~\ref{fig:motivation}(a), we inject the significance-aware binarization map into the binarization objective as an error reweighting term. \textbf{Finally}, we perform binarization fitting through alternating significance-weighted update scheme. As depicted in Fig.~\ref{fig:motivation}(c), extensive experiments on representative LVLM benchmarks demonstrate that the proposed SAB-LVLM consistently outperforms existing binary PTQ approaches under an approximately 1-bit constraint, achieving superior performance on downstream tasks while preserving compression efficiency.

\begin{itemize}
% \vspace{-6mm}
\item We propose a novel Significance-Aware Binarization for Large Vision-Language Models (SAB-LVLM). To the best of our knowledge, this work represents an early exploration of weight binarization for LVLMs. 

\item We design a spatial significance map to identify weights activated under a single modality from those activated across modalities, and devise a modality-guided integration to obtain the significance-aware binarization map.

\item We develop an alternating significance-weighted update scheme to perform binarization fitting. Experiments on representative LVLM benchmarks verify that our SAB-LVLM outperforms existing binary PTQ approaches. 

\end{itemize}

% \vspace{-3mm}
\section{Related Work}

\subsection{Post-Training Quantization}
Post-training quantization (PTQ) achieves significant reductions in model storage overhead, GPU memory consumption, and inference costs by performing low-bit discretization directly on a pre-trained model without training~\cite{li2021brecq,wang2026sliderquant,SmoothQuant}. Compared to quantization-aware training (QAT)~\cite{xu2024qalora,Hao2024QuantizedPF}, PTQ eliminates the need for costly retraining processes, making it particularly well-suited for deploying large-parameter language models and vision-language models~\cite{qwt}. The primary objective of PTQ is to minimize quantization error without retraining~\cite{Yuan2023RPTQRP, Wei2023OutlierSA, Kim2023SqueezeLLMDQ}. 
BRECQ~\cite{li2021brecq} advances PTQ to lower bits through block-level reconstruction. ZeroQuant~\cite{yao2022zeroquant} proposes an efficient and economical PTQ workflow for large-scale Transformers.
GPTQ~\cite{frantar-gptq} employs layer-wise quantization using approximate second-order information, becoming a representative method for LLMs. SmoothQuant~\cite{SmoothQuant} enhances 8-bit quantization stability by smoothly migrating activation outliers to the weight side. 
Recent works~\cite{awq,owq,SpQR,Dong2024STBLLMBT} push PTQ toward ultra-low-bit quantization, particularly binary approaches. 
PB-LLM~\cite{yuan2024pbllm} showed that naive binarization severely harms LLMs and that a small set of salient weights should be preserved at higher precision. 
BiLLM~\cite{huang2024billm} systematically introduces 1-bit PTQ into LLMs through structured significant weight selection and binary residual approximation. ARB-LLM~\cite{li2025arbllm} proposed alternating refined binarization to progressively update binarization parameters and reduce the distribution gap between binarized and full-precision weights. 
However, existing binary PTQ methods primarily focus on LLMs~\cite{chen2025hbllm}, while relevant research on LVLMs remains scarce.

\vspace{-2mm}
\subsection{Large Vision-Language Models}
Large Vision-Language Models (LVLMs) extend LLMs~\cite{radford2019language,deepseekai2025deepseekv3technicalreport,qwen3technicalreport,Qwen-VL,WM_Survey,wang2026lifelong} to visual understanding by coupling a vision encoder with a large language model through lightweight alignment modules. 
Early efforts~\cite{blip,dai2023instructblip,llava} mainly focused on bridging a pretrained vision encoder and a pretrained LLM through lightweight cross-modal connectors and visual instruction tuning, so that language models could accept visual inputs and perform general image-conditioned reasoning. The Qwen family~\cite{qwen2.5-VL,qwen3technicalreport}, as powerful open-source LLMs, have been widely adopted as the backbone of LVLMs. BLIP-2~\cite{blip} uses Q-Former to connect frozen visual encoders and frozen LLMs. InstructBLIP~\cite{dai2023instructblip} builds upon this by introducing instruction-aware visual queries. LLaVA~\cite{llava} advances this approach toward more general multimodal dialogue and reasoning through large-scale visual instruction fine-tuning.
In recent years, modern LVLMs~\cite{qwen2.5-VL,wang2025internvl3_5} have acquired powerful perception and reasoning capabilities through extensive training, and have achieved success in numerous downstream tasks, such as visual question answering (VQA)~\cite{docvqa}, multimodal dialogue~\cite{das2017visualdialog}, visual grounding~\cite{Yu2016ModelingCI}, and embodied interaction~\cite{kim2024realfred}. These downstream tasks require both perceptual and linguistic reasoning abilities. However, this multimodal capability also incurs substantial deployment costs. To address this limitation, a common approach involves compressing models through paradigms such as distillation~\cite{gu2024minillm}, PTQ~\cite{bulat2024qbb}, and QAT~\cite{llmqat}. Among these, PTQ typically offers lower training costs and better plug-and-play capabilities, yet it still struggles to be deployed on edge devices. 
Binarization advances PTQ to the approximate 1-bit precision, thereby further reducing storage requirements and computational overhead.

\vspace{-2mm}
\section{Methodology}
\label{sec:method}

\begin{figure*}[t]
    \centering
    \includegraphics[width=\linewidth]{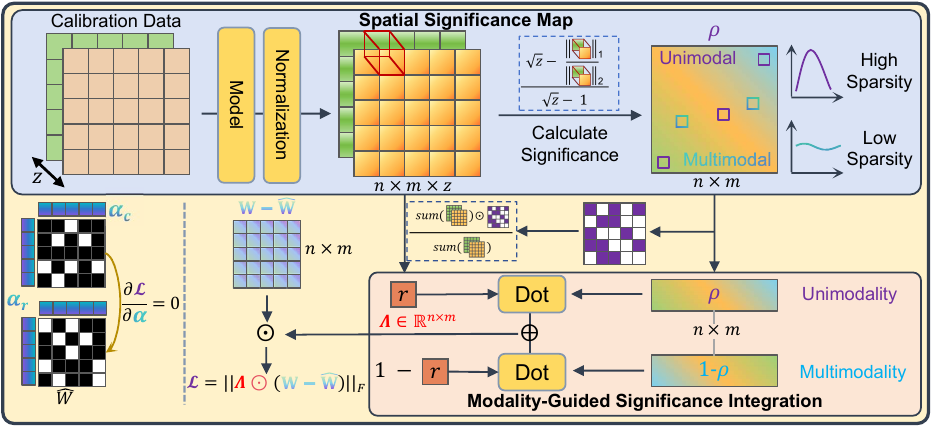}
    \vspace{-7mm}
    \caption{Overview of the proposed SAB-LVLM. The upper details the calculation process for the Spatial Significance Map: Calibrated data from different modalities are input into the model separately to compute their respective sensitivities, followed by the calculation of the Spatial Significance Map. The lower describes Modality-Guided Significance Integration and Alternating Significance-Weighted Update.}
    \label{fig:SAB_pipeline}
    \vspace{-5mm}
\end{figure*}

\subsection{Preliminary and Overview}
\label{sec:prelim}
\textbf{Preliminary:} Binarization~\cite{huang2024billm,li2025arbllm} in large vision-language models (LVLMs) compresses continuous weights into binary values (\emph{e.g.}, $\pm 1$), thereby reducing storage requirements and computational overhead. Given a full-precision weight $\mathbf{W} \in \mathbb{R}^{n \times m}$ in the original LVLMs, we define its binarization objective as:
\vspace{-2mm}
\begin{align}
\mathcal{L}(\mathbf{W}, \widehat{\mathbf{W}}) = \|\mathbf{W}-\widehat{\mathbf{W}}\|_F^2,
\label{eq:binarization_objective}
\end{align}
where $n$ and $m$ denote the row and column dimensions of $\mathbf{W}$. $\widehat{\mathbf{W}} \in \mathbb{R}^{n \times m}$ represents the low-bit weight, and it is approximated as $1$-bit in this paper. To perform binarization, we follow \cite{li2025arbllm} and represent $\widehat{\mathbf{W}}$ using a $T$-order binary expansion strategy:
\vspace{-3mm}
\begin{align}
\widehat{\mathbf{W}} =
\sum_{\epsilon=1}^{T}
\big(\boldsymbol{\alpha}^{\epsilon}_{r} (\boldsymbol{\alpha}^{\epsilon}_{c})^\top\big)
\odot \mathbf{B}^{\epsilon},
\label{eq:binary_expansion}
\end{align}
where $\mathbf{B}^{\epsilon}\in\{+1,-1\}^{n\times m}$ is the binary basis at order $\epsilon$, $\boldsymbol{\alpha}^{\epsilon}_{r}\in\mathbb{R}^{n}$ and
$\boldsymbol{\alpha}^{\epsilon}_{c}\in\mathbb{R}^{m}$ denote the row-wise and column-wise scaling vectors at order $\epsilon$. $\odot$ indicates the Hadamard product. Generally, most existing binarization methods \cite{yuan2024pbllm,chen2025hbllm,bulat2024qbb} focus solely on minimizing the weight quantization error $\mathcal{L}(\mathbf{W}, \widehat{\mathbf{W}})$ in Eq.~\eqref{eq:binarization_objective} after post-training. They overlook the fact that model weights from different layers and modalities contribute differently to downstream tasks. Evidently, this phenomenon may lead to the mistaken retention of task-irrelevant encoding parameters, resulting in significant performance degradation.

\textbf{Overview:} 
To address the above challenge, we propose a significance-aware binarization map $\boldsymbol{\Lambda} \in\mathbb{R}^{n \times m}$ to evaluate the contribution of weights across layers and modalities during binarization. $\boldsymbol{\Lambda}$ forces the optimization to preserve important weights based on their significance, enhancing the representational capacity of the binarized low-bit weights. 
Thus, we utilize $\sqrt{\boldsymbol{\Lambda}}$ to reweight Eq.~\eqref{eq:binarization_objective}: 
\begin{align}
\mathcal{L}(\mathbf{W}, \widehat{\mathbf{W}}) = \|\sqrt{\boldsymbol{\Lambda}} \odot (\mathbf{W}-\widehat{\mathbf{W}}) \|^2_F. 
\label{eq:weighted_binarization_objective}
\end{align}
As shown in Fig.~\ref{fig:SAB_pipeline}, we present the algorithmic pipeline of our model for obtaining $\boldsymbol{\Lambda}$ during post-training. 
First, we evaluate the sensitivity of model weights to different modalities based on activation values for text and image inputs, respectively. We then characterize the modal preference of local weights by computing the sparsity of sensitivity of different modalities. Second, we further compute the global preference of the weights. Finally, by integrating local and global preference, we compute dynamic sensitivity sparsity factor $\sqrt{\boldsymbol{\Lambda}}$ to perform weighting to the Eq.~\eqref{eq:binarization_objective} and then optimize $\mathbf{\widehat{W}}$ through an alternating iteration process. 

\begin{algorithm}[t]
\caption{Pipeline of The Proposed SAB-LVLM}
\label{alg:mwarb}
\begin{minipage}[t]{0.48\textwidth}
\raggedright
\textbf{Input:} Full-precision weights $\mathbf{W} \in \mathbb{R}^{n \times m}$, calibration data $\mathbf{X}$, binarization order $T$, damping coefficient $\eta$, sparsity threshold $\tau$, and iterations $N$; \\
\textbf{Output:} Low-bit weights $\widehat{\mathbf{W}} \in \mathbb{R}^{n \times m}$.\\
\textcolor{blue}{$\triangleright$ \textbf{Significance-Aware Binarization Map (sec. \ref{sec:modality}) }}\\
\begin{algorithmic}[1]
\STATE Obtain $\{\mathbf{H}^t, \mathbf{H}^v\}$ via Eq.~\eqref{eq:Hessian_matrices};
\STATE Obtain $\mathcal{S}$ via Eq.~\eqref{eq:sensitivity};
\STATE Obtain $\mathbf{\Gamma}$ via Eq.~\eqref{eq:hoyer};
\STATE Compute $r$ via Eq.~\eqref{eq:mask-uni};
\STATE Output $\mathbf{\Lambda}$ via Eq.~\eqref{eq:lambda};
\end{algorithmic}
\end{minipage}%
\hfill
\begin{minipage}[t]{0.48\textwidth}
\raggedright
% \COMMENT{\textcolor{blue}{Alternating Significance-Weighted Update})
\textcolor{blue}{$\triangleright$ \textbf{Alternating Significance-Weighted Update (sec. \ref{sec:weighted-arb}) }}  \\
% \vspace{-1mm}
\begin{algorithmic}[1]
\setcounter{ALC@line}{9}
% \item[] \textcolor{blue}{Alternating Significance-Weighted Update}
\FOR{$\epsilon=1,\dots,T$}
    \STATE Obtain $\widetilde{\mathbf{W}}=\mathbf{W}-\widehat{\mathbf{W}}$; \STATE Obtain $\mathbf{B}^{\epsilon}=\mathrm{sign}(\widetilde{\mathbf{W}})$;
    \STATE Follow \cite{li2025arbllm} to initialize $\{\boldsymbol{\alpha}_{r}^{\epsilon}, \boldsymbol{\alpha}_{c}^{\epsilon}\}$; 
    \FOR{$n=1,2,\dots,N$}
        \STATE Obtain $\alpha^{\epsilon}_{r}[i]$ via Eq.~\eqref{eq:alpha-r};
        \STATE Obtain $\alpha^{\epsilon}_{c}[j]$ via Eq.~\eqref{eq:alpha-c};
    \ENDFOR
    \STATE $\widehat{\mathbf{W}}\leftarrow \widehat{\mathbf{W}}+
    \big(\boldsymbol{\alpha}_{r}^{\epsilon}(\boldsymbol{\alpha}_{c}^{\epsilon})^\top\big)
    \odot\mathbf{B}^{\epsilon}\odot\mathcal{M}^{g}$;
\ENDFOR
% \UNTIL{convergence}
\STATE \textbf{Return} $\widehat{\mathbf{W}}$.
\end{algorithmic}
\end{minipage}
\end{algorithm}
% \vspace{-2mm}

\subsection{Significance-Aware Binarization Map}
\label{sec:modality}
As introduced in \cite{dong2019hawq,dettmers2022gpt3_int8}, sensitivity plays an important role in determining weight saliency during binarization. Inspired by these works, we utilize the Hessian matrix to quantify the sensitivity of weights with respect to different modalities for the significance-aware binarization map $\boldsymbol{\Lambda}$. 
Specifically, we first sample $K$ pairs of samples $\mathbf{X} = \{\mathbf{x}_k^t, \mathbf{x}_k^v\}_{k=1}^K$ from COCO 2017 \cite{lin2014microsoft_coco} as calibration data, where $\mathbf{x}_k^t$ and $\mathbf{x}_k^v$ represent the textual and visual modalities of the $k$-th sample. Subsequently, we input the $k$-th calibration sample $\{\mathbf{x}_k^t, \mathbf{x}_k^v\}$ into LVLM to obtain $\{\mathbf{E}_k^t, \mathbf{E}_k^v\}$. Here, $\mathbf{E}_k^t \in\mathbb{R}^{l_p\times m}$ and $\mathbf{E}_k^v \in\mathbb{R}^{l_p\times m}$ denote the hidden states of textual and visual modalities, and $l_p$ is the token length. After constructing the Hessian matrices $\mathbf{H}_k^t = 2(\mathbf{E}_k^t)^\top \mathbf{E}_k^t \in \mathbb{R}^{m \times m}$ and $\mathbf{H}_k^v = 2(\mathbf{E}_k^v)^\top \mathbf{E}_k^v \in \mathbb{R}^{m \times m}$ for the $k$-th sample, we aggregate them over all samples to obtain $\mathbf{H}^t, \mathbf{H}^v \in \mathbb{R}^{m \times m}$: 
\begin{align}
\mathbf{H}^t=\sum^K_{k=1}\mathbf{H}_k^t + \eta^t \mathbf{I},~ \mathbf{H}^v=\sum^K_{k=1}\mathbf{H}_k^v + \eta^v \mathbf{I},
\label{eq:Hessian_matrices}
\end{align}
where $I$ is the identity matrix, and $\{\eta^t, \eta^v\}$ denote the damping coefficients used to ensure numerical stability. In this paper, we set $\eta^t=\mathrm{Tr}(\sum^K_{k=1}\mathbf{H}_k^t)/m$ and $\eta^v=\mathrm{Tr}(\sum^K_{k=1}\mathbf{H}_k^v)/m$. Then, we employ Hessian matrices to compute sensitivity matrix $\mathcal{S} \in\mathbb{R}^{n\times m\times z}$, and $\mathcal{S}_{ij} \in\mathbb{R}^z$ is defined as the $(i,j)$-th element of $\mathcal{S}$:
\vspace{-2mm}
\begin{align}
\mathcal{S}_{ij} = \mathrm{Concat}(\mathcal{S}_{ij}^t, \mathcal{S}_{ij}^v),
% ~\mathcal{S}_{ij}^t = \mathbf{W}_{ij}^2/(\mathbf{H}^{t}_{jj})^{-2},~ \mathcal{S}_{ij}^v = \mathbf{W}_{ij}^2/(\mathbf{H}^{v}_{jj})^{-2},
\label{eq:sensitivity}
\end{align}
\vspace{-5mm}
\begin{align}
% \mathcal{S}_{ij} = \mathrm{Concat}(\mathcal{S}_{ij}^t, \mathcal{S}_{ij}^v),~
\mathcal{S}_{ij}^t = \mathbf{W}_{ij}^2/(\mathbf{H}^{t}_{jj})^{-2},~ \mathcal{S}_{ij}^v = \mathbf{W}_{ij}^2/(\mathbf{H}^{v}_{jj})^{-2},
\label{eq:sensitivity_component}
\end{align}

where $z$ is the number of modalities, and we set $z=2$ for textual and visual modalities. $\mathrm{Concat}(\cdot)$ denotes the concatenation operation. $\mathbf{W}_{ij}$ is the $(i, j)$-th element of the weight $\mathbf{W}$, $\mathbf{H}^{t}_{jj}$ and $\mathbf{H}^{v}_{jj}$ are the $(j, j)$-th entry of $\mathbf{H}^{t}$ and $\mathbf{H}^{v}$.

\subsubsection{Spatial Significance Map.} 
To evaluate the significance of the full-precision weight $\mathbf{W}$ at different spatial encoding locations, we propose to construct a spatial significance map $\mathbf{\Gamma} \in\mathbb{R}^{n\times m}$ by measuring the sparsity of $\mathcal{S}$. As a result, the $(i, j)$-th element $\mathbf{\Gamma}_{ij}$ of $\mathbf{\Gamma}$ can be formulated as follows: 
\begin{align}
\label{eq:hoyer}
\mathbf{\Gamma}_{ij}
&=
\frac{\sqrt{z}-\frac{\|\mathcal{S}_{ij}\|_1}{\|\mathcal{S}_{ij}\|_2}}
{\sqrt{z}-1}\in(0, 1),~
% \|\mathcal{S}_{ij}\|_1 = \sum_{u=1}^{z} \left|\mathcal{S}^{u}_{ij}\right|,~
% \|\mathcal{S}_{ij}\|_2 = \left(\sum_{u=1}^{z} (\mathcal{S}^{u}_{ij})^2\right)^{1/2},
\end{align}
where $\|\mathcal{S}_{ij}\|_1 = \sum_{u=1}^{z} \left|\mathcal{S}^{u}_{ij}\right|$ and $\|\mathcal{S}_{ij}\|_2= \left(\sum_{u=1}^{z} (\mathcal{S}^{u}_{ij})^2\right)^{1/2}$ denote the $\ell_1$ and $\ell_2$ norms of $\mathcal{S}_{ij}$. 
As can be seen from Eq.~\eqref{eq:hoyer}, when the $(i,j)$-th weight $\mathbf{W}_{ij}$ is activated only by inputs from a single modality, $\|\mathcal{S}_{ij}\|_1/\|\mathcal{S}_{ij}\|_2 \to 1$, corresponding to $\mathbf{\Gamma}_{ij} \to 1$. Otherwise, when the $(i,j)$-th weight $\mathbf{W}_{ij}$ is activated under inputs from different modalities, $\|\mathcal{S}_{ij}\|_1/\|\mathcal{S}_{ij}\|_2 \to \sqrt{z}$, corresponding to $\mathbf{\Gamma}_{ij} \to 0$. Therefore, $\mathbf{\Gamma}_{ij}$ approaches $1$ when $\mathcal{S}_{ij}$ concentrates on a single modality (unimodality), and $\mathbf{\Gamma}_{ij}$ approaches $0$ when $\mathcal{S}_{ij}$ spreads across modalities (multimodality). 

% \vspace{-3mm}
\subsubsection{Modality-Guided Significance Integration.}
Although $\mathbf{\Gamma}$ in Eq.~\eqref{eq:hoyer} measures spatial significance of encoding weights, it inherently biases the binarization process toward single-modality information, thus neglecting the complementary interactions across multiple modalities. To address this challenge, we leverage a modality integration score $r\in\mathbb{R}$ to perform modality-guided significance integration. In this paper, we employ the sensitivity matrix $\mathcal{S}$ to compute $r$:
\begin{align}
\label{eq:mask-uni}
r = 
\frac{\sum_{i,j}(\mathcal{S}_{ij}^{t}+\mathcal{S}_{ij}^{v})\cdot\mathcal{M}_{ij}^{\text{uni}}}
{\sum_{i,j}(\mathcal{S}_{ij}^{t}+\mathcal{S}_{ij}^{v})}
\in(0,1), \;
% \mathcal{M}^{uni}_{ij}=\left\{\begin{matrix} 
%   1, \; if \; \mathbf{\Gamma}_{ij}>\tau, \\  
%   0, \; otherwise,
% \end{matrix}\right.
\end{align}
\vspace{-3mm}
\begin{align}
\label{eq:mask}
\mathcal{M}^{uni}_{ij}=\left\{\begin{matrix} 
  1, \; if \; \mathbf{\Gamma}_{ij}>\tau, \\  
  0, \; otherwise,
\end{matrix}\right.
\end{align}
where $\tau$ is the threshold for controlling sparsity. If $\mathbf{\Gamma}_{ij} > \tau$, we consider the $(i,j)$-th element $\mathbf{W}_{ij}$ tends to be activated under a single modality; otherwise, it tends to be activated under inputs from different modalities. 
Therefore, the significance-aware binarization map $\boldsymbol{\Lambda}\in\mathbb{R}^{n\times m}$ is defined as follows:
\begin{align}
\label{eq:lambda}
\boldsymbol{\Lambda}
=
\underbrace{r \cdot\mathbf{\Gamma}}_{{\text{\scriptsize Unimodal}} } \;+\;
\underbrace{(1-r)\cdot(1-\mathbf{\Gamma})}_{{\text{\scriptsize Multimodal}}}.
\end{align}
Evidently, larger values of $\boldsymbol{\Lambda}$ in Eq.~\eqref{eq:lambda} impose a greater penalty on the quantization error in Eq.~\eqref{eq:weighted_binarization_objective}. 
The significance-aware binarization map $\boldsymbol{\Lambda}$ explicitly implements an error-control mechanism: when the modality integration score $r > 0.5$, full-precision weights activated by a single modality receive larger $\boldsymbol{\Lambda}$ values; otherwise, weights activated across multiple modalities are emphasized.

\vspace{-2mm}
\begin{tcolorbox}[colback=green!3, colframe=black, boxrule=0.2 mm]
\begin{theorem}
% \vspace{-2mm}
\label{theorem1}
\textbf{Monotonicity of Quantization Error} \\
Let $e_{ij} := |\mathbf{W}_{ij}-\widehat{\mathbf{W}}_{ij}|$ denote the element-wise quantization error, which is continuous and independent.
Consider the relaxed error-allocation problem associated with Eq.~\eqref{eq:weighted_binarization_objective}:
% \vspace{-1mm}
$
\min_{\{e_{ij}\ge 0\}} \sum_{i,j}\mathbf{\Lambda}_{ij} e_{ij}^2
\;\text{s.t.}\; \sum_{i,j} e_{ij} = E,
$
where $E>0$ is a fixed total error budget.
Let $e_{ij}^*$ be the optimal quantization error. 
For any two elements $(i,j)$ and $(p,q)$, the optimal error satisfies:
% \vspace{-5mm}
\begin{align}
\label{theorem1_eq}
e_{ij}^* \le \frac{\boldsymbol{\Lambda}_{pq}}{\boldsymbol{\Lambda}_{ij}}\, e_{pq}^*.
\end{align}
% and in fact $e_{ij}^*$ is inversely proportional to $\boldsymbol{\Lambda}_{ij}$, i.e., $e_{ij}^* \propto{\boldsymbol{\Lambda}_{ij}}^{-1}$.
\end{theorem}
\vspace{-5mm}
\end{tcolorbox}

Theorem~\ref{theorem1} shows that for any two distinct elements $(i,j)$ and $(p,q)$, 
$\boldsymbol{\Lambda}_{ij} > \boldsymbol{\Lambda}_{pq}$ implies $e_{ij}^* < e_{pq}^*$. 
Moreover, the optimal error is inversely proportional to the squared significance weight, 
i.e., $e_{ij}^* \propto \boldsymbol{\Lambda}_{ij}^{-1}$. 
From a theoretical perspective, this theorem demonstrates that full-precision weights corresponding to larger values in the significance-aware binarization map $\boldsymbol{\Lambda}$ incur smaller quantization errors. Theorem \ref{theorem1} thus provides theoretical support for our claim that the proposed significance-aware binarization map $\boldsymbol{\Lambda}$ guides the optimization process to preserve task-relevant weights according to their significance.

\begin{table*}[t]
\centering
\footnotesize
\setlength{\tabcolsep}{2pt}
\renewcommand{\arraystretch}{1.15}
\caption{Performance comparison of different quantization methods on Qwen2.5-VL-Instruct models of different scales. 
The best result is highlighted in \best{red}.}
\vspace{-3mm}
\label{tab:qwen25_quant_results}
\begin{adjustbox}{max width=\textwidth}
\begin{tabular}{@{}l|l|ccccccc@{}}
\toprule
Model & Method & Weight bits & Memory & MMStar & DocVQA& TextVQA & Video-MME& VSI-Bench\\
\midrule
\multirow{6}{*}{\makecell[l]{Qwen2.5-VL-7B-\\Instruct}}
& Full Precision            & 16 bit   & 15.45 GB & 62.25 & 95.17 & 83.88 & 61.96 & 29.75 \\
\cmidrule(l){2-9}
& GPTQ~\cite{frantar-gptq} (ICLR 2023)           & 3 bit    & 2.66 GB & 56.10 & 93.36 & 82.25 & 59.93 & 33.16 \\
\cmidrule(l){2-9}
& PB-LLM~\cite{yuan2024pbllm} (ICLR'2024)        & 1.17 bit & 2.15 GB & 4.6  & 9.40  & 28.67 & 9.89  & 4.94 \\
& BiLLM~\cite{huang2024billm} (ICML'2024)        & 1.08 bit & 2.24 GB & 29.97 & 33.08 & 53.38 & 26.56 & 13.12 \\
& ARB-LLM~\cite{li2025arbllm} (ICLR'2025)       & 1.07 bit & 2.14 GB & 38.10 & 80.48 & 70.70 & 47.14 & 25.89 \\
\rowcolor{gray!15}{\cellcolor{white}}
& \textbf{Ours} (\textbf{SAB-LVLM})  & 1.07 bit & 2.14 GB & \best{45.79} & \best{85.34} & \best{74.00} & \best{48.11} & \best{26.01} \\
\midrule

\multirow{6}{*}{\makecell[l]{Qwen2.5-VL-32B-\\Instruct}}
& Full Precision            & 16 bit   & 62.31 GB & 69.50 & 94.80 & 78.85 & 70.50 & 36.99 \\
\cmidrule(l){2-9}
& GPTQ~\cite{frantar-gptq} (ICLR 2023)           & 3 bit    & 12.71 GB & 58.63    & 92.65 & 74.45 & 61.67 & 33.90 \\
\cmidrule(l){2-9}
& PB-LLM~\cite{yuan2024pbllm} (ICLR'2024)        & 1.70 bit & 10.24 GB & 37.66 & 66.10 & 63.62 & 47.89 & 15.21 \\
& BiLLM~\cite{huang2024billm} (ICML'2024)        & 1.08 bit & 10.73 GB  & 46.29 & 60.85 & 55.92 & 47.04 & 17.65 \\
& ARB-LLM (ICLR'2025)       & 1.16 bit & 10.28 GB  & 48.48 & 90.00 & 74.96 & 57.70 & 25.12 \\
\rowcolor{gray!15}{\cellcolor{white}}
& \textbf{Ours} (\textbf{SAB-LVLM})   & 1.07 bit & 10.28 GB  & \best{54.77} & \best{90.36} & \best{75.45} & \best{59.15} & \best{29.05} \\
\midrule

\multirow{6}{*}{\makecell[l]{Qwen2.5-VL-72B-\\Instruct}}
& Full Precision            & 16 bit   & 136.74 GB & 67.19 & 96.40 & 83.26 & 73.30 & 37.00 \\
\cmidrule(l){2-9}
& GPTQ~\cite{frantar-gptq} (ICLR 2023)           & 3 bit    & 28.61 GB & 66.57    & --94.25   & --82.03 & 65.11  & 34.71   \\
\cmidrule(l){2-9}
& PB-LLM~\cite{yuan2024pbllm} (ICLR'2024)        & 1.70 bit & 22.99 GB & 30.14 & 85.33 & 74.34 & 54.37 & 17.27 \\
& BiLLM~\cite{huang2024billm} (ICML'2024)        & 1.08 bit & 24.23 GB & 51.62 & 90.10 & 77.28 & 58.41 & 25.76 \\
& ARB-LLM~\cite{li2025arbllm} (ICLR'2025)        & 1.19 bit & 23.23 GB & 57.45 & 90.42 & 80.23 & 61.15 & 33.40 \\
\rowcolor{gray!15}{\cellcolor{white}}
& \textbf{Ours} (\textbf{SAB-LVLM})   & 1.07 bit & 23.23 GB & \best{58.48} & \best{92.56} & \best{80.76} & \best{64.93} & \best{34.18} \\
\bottomrule
\end{tabular}
\end{adjustbox}
% \vspace{-5mm}
\end{table*}

\begin{table*}[h]
\centering
\footnotesize
\setlength{\tabcolsep}{2pt}
\renewcommand{\arraystretch}{1.15}
\caption{Performance comparison of different quantization methods on InternVL3.5-Instruct across five multimodal benchmarks. 
% The best result is highlighted in \best{red}.
}
\vspace{-3mm}
\label{tab:internvl35_8b_main_results}
\begin{adjustbox}{max width=\textwidth}
\begin{tabular}{@{}l|l|ccccccc@{}}
\toprule
Model & Method & Weight bits & Memory & MMStar & DocVQA& TextVQA & Video-MME& VSI-Bench \\
\midrule

\multirow{6}{*}{\makecell[l]{InternVL3.5-8B-\\Instruct}}
& Full Precision      & 16 bit   & 15.89 GB & 66.15 & 92.47 & 77.38 & 66.59 & 56.30 \\
\cmidrule(l){2-9}
& GPTQ~\cite{frantar-gptq} (ICLR 2023)    & 3 bit    & 2.83 GB & 61.04 & 89.82 & 74.82 & 56.81 & 49.01 \\
\cmidrule(l){2-9}
& PB-LLM~\cite{yuan2024pbllm} (ICLR'2024)  & 1.70 bit & 2.28 GB & 10.36 & 0.03 & 0.00 & 24.67 & 0.26 \\
& BiLLM~\cite{huang2024billm} (ICML'2024)   &1.08 bit & 2.39 GB & 3.42 & 0.93 & 0.23 & 10.52 & 1.24 \\
& ARB-LLM~\cite{li2025arbllm} (ICLR'2025) & 1.07 bit& 2.28 GB & 24.67 & 40.67 & 30.83 & 9.89 & 0.16  \\
\rowcolor{gray!15}{\cellcolor{white}}
& \textbf{Ours} (\textbf{SAB-LVLM})     & 1.07 bit & 2.28 GB & \best{27.02} & \best{48.03} & \best{42.04} & 24.00 & \best{9.63} \\

\midrule
\multirow{6}{*}{\makecell[l]{InternVL3.5-14B-\\Instruct}}
& Full Precision      & 16 bit & 28.16 GB  & 65.18 & 93.40 & 77.80 & 67.90 & 60.80 \\
\cmidrule(l){2-9}
& GPTQ~\cite{frantar-gptq} (ICLR 2023)            & 3 bit & 5.38 GB & 64.45 & 94.01 & 74.90 & 60.07 & 52.60 \\
\cmidrule(l){2-9}
& PB-LLM~\cite{yuan2024pbllm} (ICLR'2024)         & 1.70 bit  & 4.34 GB & 12.53 & 0.00 & 0.00 & 22.52  & 0.09 \\
& BiLLM~\cite{huang2024billm} (ICML'2024)         & 1.07 bit  & 4.53 GB & 31.38 & 47.09 & 31.65 & 15.63  & 17.36 \\
& ARB-LLM~\cite{li2025arbllm} (ICLR'2025)         & 1.07 bit & 4.33 GB & 47.50 & 85.15 & 70.77 & 51.33 & 31.50 \\
\rowcolor{gray!15}{\cellcolor{white}}
& \textbf{Ours} (\textbf{SAB-LVLM})               & 1.07 bit  & 4.33 GB & \best{50.46} & 84.34 & \best{71.03} & \best{51.96} & \best{32.50} \\
\midrule

\multirow{6}{*}{\makecell[l]{InternVL3.5-38B-\\Instruct}}
& Full Precision      & 16 bit   & 71.51 GB & 75.30 & 94.00 & 82.70 & 70.90 & 66.30 \\
\cmidrule(l){2-9}
& GPTQ~\cite{frantar-gptq} (ICLR 2023)            & 3 bit & 12.71 GB & 68.04 & 90.25 & 80.22 & 62.30 & 56.12 \\
\cmidrule(l){2-9}
& PB-LLM~\cite{yuan2024pbllm} (ICLR'2024)         & 1.70 bit    & 10.24 GB & 37.07 & 54.02 & 65.06 & 46.85  & 26.83 \\
& BiLLM~\cite{huang2024billm} (ICML'2024)         & 1.07 bit    & 10.72 GB & 21.30 & 42.90 & 37.09 & 16.93  & 26.71 \\
& ARB-LLM~\cite{li2025arbllm} (ICLR'2025)         & 1.08 bit    & 10.27 GB & 54.54 & 86.85 & 77.37 & 59.29 & 41.62 \\
\rowcolor{gray!15}{\cellcolor{white}}
& \textbf{Ours} (\textbf{SAB-LVLM})  & 1.06 bit  & 10.27 GB & \best{56.22} & \best{87.19} & \best{78.06} & \best{59.41} & \best{43.00} \\
\bottomrule
\end{tabular}
\end{adjustbox}
\vspace{-3mm}
\end{table*}

\vspace{-3mm}
\subsection{Alternating Significance-Weighted Update}
\label{sec:weighted-arb}
Inspired by \cite{li2025arbllm, huang2024billm, frantar-gptq}, we propose an alternating refinement procedure to optimize Eq.~\eqref{eq:weighted_binarization_objective} based on a sensitivity-aware binarization map $\boldsymbol{\Lambda}$. Given the sensitivity matrix $\mathcal{S}$, we set $T=2$ for key weights to achieve enhanced representation
capabilities; otherwise, $T=1$. Additionally, in Eq. \eqref{eq:binary_expansion}, we set the binary basis $\mathbf{B}=\mathrm{sign}(\mathbf{W})$ and use $\partial \mathcal{L}(\mathbf{W}, \widehat{\mathbf{W}})/\partial \alpha^{\epsilon}_r = 0$ to update row-wise scaling vector $\alpha^{\epsilon}_{r} \in\mathbb{R}^n$ at order $\epsilon$:
\vspace{-3mm}
\begin{align}
\label{eq:alpha-r}
\alpha^{\epsilon}_{r}[i]
=
\frac{\sum_{j=1}^{m}\boldsymbol{\Lambda}_{ij}\, \mathbf{\Phi}^{\epsilon}_{ij}\, \widetilde{\mathbf{W}}_{ij}}
{\sum_{j=1}^{m}\boldsymbol{\Lambda}_{ij}\, (\mathbf{\Phi}^{\epsilon}_{ij})^2},\; 
\mathbf{\Phi}^{\epsilon}_{ij} = \alpha^{\epsilon}_{c}[j]\mathbf{B}^{\epsilon}_{ij} \mathcal{M}^g_{ij},
\end{align}
where $\widetilde{\mathbf{W}} = \mathbf{W} - \mathbf{\widehat{W}} \in \mathbb{R}^{n \times m}$ is the residual matrix, $\alpha^{\epsilon}_{r}[i]$ is the scale value in the $i$-th row, and $\alpha^{\epsilon}_{c}[j]$ denotes the scale value of the $j$-th column at order $\epsilon$. $\mathcal{M}^g$ represents the group mask generated by \cite{frantar-gptq}, and $\mathcal{M}^g_{ij}$ indicates the $(i, j)$-th element of $\mathcal{M}^g$. Similar to $\alpha^{\epsilon}_r$, we update the column-wise scaling vector $\alpha^{\epsilon}_c\in\mathbb{R}^m$ at order $\epsilon$ by setting $\partial \mathcal{L}(\mathbf{W}, \widehat{\mathbf{W}})/\partial \alpha^{\epsilon}_c=0$:
\begin{align}
\label{eq:alpha-c}
\alpha^{\epsilon}_{c}[j]
=
\frac{\sum_{i=1}^{n} \boldsymbol{\Lambda}_{ij}\mathbf{\Psi}^{\epsilon}_{ij}\, \widetilde{\mathbf{W}}_{ij}}
{\sum_{i=1}^{n} \boldsymbol{\Lambda}_{ij}(\mathbf{\Psi}^{\epsilon}_{ij})^2},
\;
\mathbf{\Psi}^{\epsilon}_{ij} = \alpha^{\epsilon}_{r}[i]\mathbf{B}^{\epsilon}_{ij}\mathcal{M}^g_{ij},
\end{align}
The optimization pipeline of our proposed SAB-LVLM is shown in Algorithm \ref{alg:mwarb}.

\section{Experiments}
\subsection{Setup}

\textbf{Implementation Details.}
All experiments were conducted on a single NVIDIA A100 GPU (80G) and evaluated using lmms-eval~\cite{zhang2024lmmsevalrealitycheckevaluation} for zero-shot testing. We set the block size to 128 and performed 15 iterations of alternating significance-weighted update. We sampled $K$ examples from the COCO 2017~\cite{lin2014microsoft_coco} dataset as calibration data. In this paper,we set $K=128$. Additionally, as a PTQ method, SAB-LVLM did not apply any training or fine-tuning of the model.
% \subsection{Models and Benchmarks}

\textbf{Models. }To comprehensively demonstrate the effectiveness of our approach, we evaluate the proposed SAB-LVLM across different LVLMs, including Qwen2.5-VL~\cite{qwen2.5-VL,Qwen-VL,Qwen2VL} and InternVL3.5~\cite{wang2025internvl3_5} families. Our evaluation spans a diverse range of model capacities, specifically targeting the 7B, 32B, and 72B variants of Qwen2.5-VL, alongside the 8B, 14B, and 38B variants of InternVL3.5. 

% \begin{figure*}[t]
%     \centering
%     \includegraphics[width=\textwidth]{Figures/mmstar_vis_0.pdf}
%     \vspace{-5mm}
%     \caption{Qualitative analysis at MMStar with Qwen2.5-VL-7B-Instruct.}
%     \label{mmstar_vis}
%     \vspace{-3mm}
% \end{figure*}

\textbf{Benchmarks. }
Furthermore, to rigorously assess the versatility of SAB-LVLM across various downstream tasks, we conduct extensive experiments on multiple benchmarks, including MMStar~\cite{chen2024mmstar}, DocVQA~\cite{docvqa}, TextVQA~\cite{textvqa}, Video-MME~\cite{videomme}, and VSI-Bench~\cite{Vsibench}.
MMStar~\cite{chen2024mmstar} is a comprehensive benchmark for LVLMs, comprising 1,500 meticulously curated samples designed to assess model capabilities across six core competencies. DocVQA~\cite{docvqa} is a document-visual question-answering dataset comprising 50,000 questions based on over 12,000 document images. It primarily evaluates ability to understand and reason about document content and layout structures. TextVQA~\cite{textvqa} is a visual question-answering dataset that requires LVLMs to read and understand text within natural scene images, then answer questions by integrating visual context. Video-MME~\cite{videomme} is a comprehensive evaluation benchmark for assessing the video analysis and temporal understanding capabilities of LVLMs. VSI-Bench~\cite{Vsibench} is a video benchmark for evaluating the visual-spatial intelligence of LVLMs. It constructs over 5,000 question-answer pairs, focusing on assessing spatial relationship comprehension, metric estimation, and spatio-temporal reasoning.

\begin{table*}[t]
\centering
\footnotesize
\setlength{\tabcolsep}{2pt}
\renewcommand{\arraystretch}{1.15}
\caption{Detailed Results on Qwen2.5-VL-Instruct at MMStar. 
Coa. Prcep.: Coarse Perception; Fin. Prcep.: Fine-grained Perception; Ins. Reas.: Instance Reasoning; Logic. Reas.: Logical Reasoning; Math.: Mathematics; Sci. \& Tech.: Science \& Technology. 
% The best result is highlighted in \textcolor{red}{red}.
}
\vspace{-3mm}
\label{tab:qwen25_mmstar_results}
\begin{adjustbox}{max width=\textwidth}
\begin{tabular}{@{}l|l|cccccccc@{}}
\toprule
Model & Method & Weight bits & Memory & Coa. Prcep. & Fin. Prcep. & Ins. Reas.& Logic. Reas. & Math. & Sci. \& Tech. \\
\midrule

\multirow{6}{*}{\makecell[l]{Qwen2.5-7B-\\Instruct}}
& Full Precision       & 16 bit   & 15.45 GB  & 90.74 & 61.13 & 72.30 & 61.64 & 59.26 & 38.33 \\
\cmidrule(l){2-10}
& GPTQ~\cite{frantar-gptq} (ICLR 2023)     & 3 bit    & 2.66 GB  & 70.93 & 53.77 & 65.34 & 55.85 & 52.87 & 37.83 \\
\cmidrule(l){2-10}
& PB-LLM~\cite{yuan2024pbllm} (ICLR'2024)  & 1.17 bit & 2.15 GB  & 2.81 & 0.64 & 2.77 & 5.92 & 9.92 & 5.53  \\
& BiLLM~\cite{huang2024billm} (ICML'2024)  & 1.08 bit & 2.24 GB  & 46.56 & 26.10 & 35.46 & 25.00 & 22.08 & 24.65 \\
& ARB-LLM~\cite{li2025arbllm} (ICLR'2025)  & 1.07 bit & 2.14 GB  & 62.97 & 31.52 & 45.93 & 41.91 & 26.74 & 19.54 \\
\rowcolor{gray!15}{\cellcolor{white}}
& \textbf{Ours} (\textbf{SAB-LVLM})        & 1.07 bit & 2.14 GB  & \best{71.49} & \best{39.12} & \best{53.35} & \best{42.71} & \best{38.17} & \best{29.88} \\
\midrule

\multirow{6}{*}{\makecell[l]{Qwen2.5-32B-\\Instruct}}
& Full Precision       & 16 bit   & 62.31 GB  & 75.11 & 57.06 & 72.91 & 69.32 & 75.17 & 51.57 \\
\cmidrule(l){2-10}
& GPTQ~\cite{frantar-gptq} (ICLR 2023)     & 3 bit    & 12.71 GB & 71.85 & 54.53 & 70.71 & 57.61 & 58.36 & 38.72 \\
\cmidrule(l){2-10}
& PB-LLM~\cite{yuan2024pbllm} (ICLR'2024)  & 1.70 bit & 10.24 GB & 59.11 & 34.76 & 51.22 & 32.93 & 27.78 & 20.15 \\
& BiLLM~\cite{huang2024billm} (ICML'2024)  & 1.08 bit & 10.73 GB  & 69.77 & 43.68 & 56.85 & 43.15 & 37.53 & 26.76 \\
& ARB-LLM~\cite{li2025arbllm} (ICLR'2025)  & 1.16 bit & 10.28 GB  & 69.78 & 47.81 & 61.46 & 44.86 & 33.77 & 33.21 \\
\rowcolor{gray!15}{\cellcolor{white}}
& \textbf{Ours} (\textbf{SAB-LVLM})        & 1.07 bit & 10.28 GB  & \best{71.17} & 43.89 & \best{67.40} & \best{57.31} & \best{52.11} & \best{36.73} \\
\midrule

\multirow{6}{*}{\makecell[l]{Qwen2.5-72B-\\Instruct}}
& Full Precision       & 16 bit   & 136.74 GB & 75.86 & 60.30 & 72.08 & 73.94 & 71.74 & 48.16 \\
\cmidrule(l){2-10}
& GPTQ~\cite{frantar-gptq} (ICLR'2023)     & 3 bit    & 28.61 GB & 76.08 & 60.56 & 70.70 & 74.50 & 69.33 & 48.26 \\
\cmidrule(l){2-10}
& PB-LLM~\cite{yuan2024pbllm} (ICLR'2024)  & 1.70 bit & 22.99 GB & 60.22 & 37.69 & 46.69 & 21.25 & 7.80  & 7.21 \\
& BiLLM~\cite{huang2024billm} (ICML'2024)  & 1.08 bit & 24.23 GB & 67.94 & 49.51 & 62.74 & 49.18 & 51.53 & 28.81 \\
& ARB-LLM~\cite{li2025arbllm} (ICLR'2025)  & 1.19 bit & 23.23 GB & 73.03 & 53.70 & 66.09 & 55.44 & 60.99 & 35.46 \\
\rowcolor{gray!15}{\cellcolor{white}}
& \textbf{Ours} (\textbf{SAB-LVLM})        & 1.07 bit & 23.23 GB & 72.57 & \best{56.37} & 65.57 & \best{60.81} & \best{62.06} & 33.51 \\
\bottomrule
\end{tabular}
\end{adjustbox}
\vspace{-5mm}
\end{table*}

\begin{table*}
\centering
\footnotesize
\setlength{\tabcolsep}{2pt}
\renewcommand{\arraystretch}{1.15}
\caption{Detailed Results on InternVL3.5 Instruct at MMStar. 
Coa. Prcep.: Coarse Perception; Fin. Prcep.: Fine-grained Perception; Ins. Reas.: Instance Reasoning; Logic. Reas.: Logical Reasoning; Math.: Mathematics; Sci. \& Tech.: Science \& Technology. 
% The best result is highlighted in \textcolor{red}{red}.
}
\vspace{-3mm}
\label{tab:qinternvl35_mmstar_results}
\begin{adjustbox}{max width=\textwidth}
\begin{tabular}{@{}l|l|cccccccc@{}}
\toprule
Model & Method & Weight bits & Memory & Coa. Prcep. & Fin. Prcep. & Ins. Reas.& Logic. Reas. & Math. & Sci. \& Tech. \\
\midrule

\multirow{6}{*}{\makecell[l]{InternVL3.5-8B-\\Instruct}}
& Full Precision       & 16 bit   & 15.89 GB & 77.44 & 57.90 & 75.13 & 67.11 & 70.55 & 48.75 \\
\cmidrule(l){2-10}
& GPTQ~\cite{frantar-gptq} (ICLR'2023)     & 3 bit    & 2.83 GB  & 76.02 & 53.86 & 68.24 & 64.71 & 65.63 & 37.77 \\
\cmidrule(l){2-10}
& PB-LLM~\cite{yuan2024pbllm} (ICLR'2024)  & 1.70 bit & 2.28 GB  & 8.04 & 10.62 & 9.13 & 11.13 & 14.09 & 9.16 \\
& BiLLM~\cite{huang2024billm} (ICML'2024)  & 1.08 bit & 2.39 GB  & 2.17 & 5.64 & 1.96 & 4.52 & 3.13 & 3.09 \\
& ARB-LLM~\cite{li2025arbllm} (ICLR'2025)  & 1.07 bit & 2.28 GB  & 39.00 & 28.11 & 32.11 & 24.07 & 12.33 & 12.40 \\
\rowcolor{gray!15}{\cellcolor{white}}
& \textbf{Ours} (\textbf{SAB-LVLM}) & 1.07 bit & 2.28 GB  & \best{53.33} & \best{32.68} & 31.83 & 23.13 & 6.25 & \best{14.91} \\
\midrule

\multirow{6}{*}{\makecell[l]{InternVL3.5-14B-\\Instruct}}
& Full Precision       & 16 bit   & 28.16 GB & 74.41 & 58.87 & 78.54 & 72.96 & 80.29 & 56.32 \\
\cmidrule(l){2-10}
& GPTQ~\cite{frantar-gptq} (ICLR'2023)     & 3 bit    & 5.38 GB  & 76.19 & 59.65 & 69.91 & 64.72 & 67.57 & 48.67 \\
\cmidrule(l){2-10}
& PB-LLM~\cite{yuan2024pbllm} (ICLR'2024)  & 1.70 bit & 4.34 GB  & 12.30 & 7.66 & 11.39 & 15.67 & 12.76 & 15.39 \\
& BiLLM~\cite{huang2024billm} (ICML'2024)  & 1.07 bit & 4.53 GB  & 60.34 & 39.17 & 32.63 & 21.97 & 11.15 & 23.05 \\
& ARB-LLM~\cite{li2025arbllm} (ICLR'2025)  & 1.07 bit & 4.33 GB  & 64.47 & 46.55 & 54.74 & 40.98 & 48.03 & 30.23 \\
\rowcolor{gray!15}{\cellcolor{white}}
& \textbf{Ours} (\textbf{SAB-LVLM})        & 1.07 bit & 4.33 GB  & \best{67.33} & \best{50.46} & \best{57.51} & \best{43.76} & \best{52.82} & \best{30.88} \\
\midrule

\multirow{6}{*}{\makecell[l]{InternVL3.5-38B-\\Instruct}}
& Full Precision       & 16 bit   & 71.51 GB & 78.23 & 57.53 & 80.64 & 80.33 & 85.25 & 69.81 \\
\cmidrule(l){2-10}
& GPTQ~\cite{frantar-gptq} (ICLR'2023)     & 3 bit    & 12.71 GB & 74.40 & 64.54 & 72.53 & 70.81 & 74.04 & 51.96 \\
\cmidrule(l){2-10}
& PB-LLM~\cite{yuan2024pbllm} (ICLR'2024)  & 1.70 bit & 10.24 GB & 67.71 & 36.38 & 48.90 & 25.61 & 20.17 & 23.64 \\
& BiLLM~\cite{huang2024billm} (ICML'2024)  & 1.07 bit & 10.72 GB & 31.86 & 14.03 & 24.62 & 22.66 & 17.25 & 17.36 \\
& ARB-LLM~\cite{li2025arbllm} (ICLR'2025)  & 1.08 bit & 10.27 GB & 72.24 & 56.57 & 64.89 & 49.18 & 50.67 & 33.68 \\
\rowcolor{gray!15}{\cellcolor{white}}
& \textbf{Ours} (\textbf{SAB-LVLM})                & 1.06 bit & 10.27 GB & \best{75.02} & \best{58.38} & 63.54 & 47.22 & \best{53.01} & \best{40.17} \\
\bottomrule
\end{tabular}
\end{adjustbox}
\vspace{-3mm}
\end{table*}

\begin{table*}[t]
\centering
\footnotesize
\setlength{\tabcolsep}{5pt}
\renewcommand{\arraystretch}{1.12}
% \vspace{-5mm}
\caption{Detailed results on Qwen2.5-VL-32B-Instruct at VSI-Bench. 
OAO: object appearance order; OAD: object absolute distance; 
OC: object counting; ORD: object relative distance; 
OSE: object size estimation; RSE: room size estimation; 
RP: route planning; ODir: object relative direction. 
% All results are reported in percentage (\%).
}
\vspace{-3mm}
\label{tab:vsibench_detailed_qwen}
\begin{adjustbox}{max width=\textwidth}
\begin{tabular}{@{}l|ccccccccccc@{}}
\toprule
Method & Weight bits & Memory & OAO & OAD & OC & ORD & OSE & RSE & RP & ODir & Avg.  \\
\midrule
Full Precision      & 16 bit & 62.31 GB & 25.24 & 26.40 & 34.81 & 44.79 & 51.77 & 43.54 & 31.44 & 37.88 & 36.99 \\
\cmidrule(l){1-12}
% GPTQ~\cite{frantar-gptq} (ICLR'2023)     & 3 bit & 16.38 GB  & 22.14 & 24.03 & 30.44 & 40.18 & 48.89 & 39.33 & 27.52 & 33.71 \\
GPTQ~\cite{frantar-gptq} (ICLR'2023)     & 3 bit & 12.71 GB  & 25.89 & 29.92 & 28.83 & 38.03 & 43.80 & 29.55 & 32.47 & 42.74 & 33.90 \\
\cmidrule(l){1-12}
PB-LLM~\cite{yuan2024pbllm} (ICLR'2024)  &  1.70 bit  & 10.24 GB & 19.74 & 0 & 0 & 32.96 & 0 & 0 & 29.90 & 39.12 & 15.21 \\
BiLLM~\cite{huang2024billm} (ICML'2024)  &  1.08 bit  & 10.73 GB & 21.04 & 12.07 & 6.85 & 27.04 & 4.89 & 1.74 & 29.38 & 38.18 & 17.65 \\
ARB-LLM~\cite{li2025arbllm} (ICLR'2025)  &  1.07 bit  & 10.28 GB & 25.89 & 24.14 & 37.47 & 40.85 & 3.42 & 8.33 & 28.35 & 32.53 & 25.12 \\
\rowcolor{gray!15}
\textbf{Ours} (\textbf{SAB-LVLM}) & 1.07 bit  & 10.28 GB & \best{25.89} & \best{28.99} & 27.04 & 40.28 & \best{21.09} & \best{21.49} & \best{33.51} & 34.17 & \best{29.05} \\
\bottomrule
\end{tabular}
\end{adjustbox}
\vspace{-5mm}
\end{table*}

\textbf{Baselines.}
We compare with various low-bit methods~\cite{frantar-gptq,yuan2024pbllm,huang2024billm,li2025arbllm} across multiple datasets and across multiple LVLMs. GPTQ~\cite{frantar-gptq} maintains performance at lower bit widths through layer-by-layer weighting based on second-order information. Note that GPTQ~\cite{frantar-gptq} is evaluated at 3-bit quantization, and we report its results only for performance reference rather than as a 1-bit baseline. BiLLM~\cite{huang2024billm} is a 1-bit PTQ method that reduces compression error by identifying significant weights and combining them with binary residual approximation. PB-LLM~\cite{yuan2024pbllm} is a partial binarization method that preserves a small number of significant weights as high-bit representations while binarizing the remaining weights, thereby maintaining the reasoning capabilities with extremely low bit. ARB-LLM~\cite{li2025arbllm} reduces distribution bias between full-precision weights and binary weights by alternately optimizing the binarization parameters.

\subsection{Comparison Results}
To comprehensively evaluate the performance of the proposed SAB-LVLM across various downstream tasks, as shown in Tabs.~\ref{tab:qwen25_quant_results}-\ref{tab:vsibench_detailed_qwen}, we conducted a thorough assessment of the proposed SAB-LVLM against baseline methods across five multimodal benchmarks and multiple LVLMs.

\begin{table}[t]
% \vspace{-10pt}
\centering
\footnotesize
\setlength{\tabcolsep}{4pt}
\renewcommand{\arraystretch}{1.08}
\caption{Ablation study of the threshold $\tau$ on Qwen2.5-VL-32B-Instruct. Coa. Prcep.: Coarse Perception; Fin. Prcep.: Fine-grained Perception; Ins. Reas.: Instance Reasoning; Logic. Reas.: Logical Reasoning; Math.: Mathematics; Sci. \& Tech.: Science \& Technology. }
\vspace{-3mm}
\label{tab:ablation_tau_mmstar}
\begin{adjustbox}{max width=\columnwidth}
\begin{tabular}{@{}cccccccc@{}}
\toprule
% \multirow{2}{*}{$\tau$} & \multicolumn{7}{c}{MMStar} \\
% \cmidrule(lr){2-8}
$\tau$ & Coa. Prcep. & Fin. Prcep. & Ins. Reas. & Logic. Reas. & Math. & Sci. \& Tech. & Avg. \\
\midrule
Baseline & 69.78 & 47.81 & 61.46 & 44.86 & 33.77 & 33.21 & 48.48 \\
0.0001   & 69.37 & 43.45 & 61.63 & 47.25 & 36.20  & 27.87   & 47.63 \\
0.0005   & 66.51 & 42.90    & 61.48   & 49.35   & 45.47  & 30.00  & 49.28 \\
0.0003   & \best{71.17} & 43.89 & \best{67.40} & \best{57.31} & \best{52.11} & \best{36.73} & \best{54.77} \\
\bottomrule
\end{tabular}
\end{adjustbox}
\vspace{-3mm}
\end{table}

\begin{table}[t]
\vspace{-5pt}
\centering
\footnotesize
\setlength{\tabcolsep}{3pt}
\renewcommand{\arraystretch}{1.08}
\caption{Comparison of different variants of MGSI on Qwen2.5-VL-32B at MMStar. Coa. Prcep.: Coarse Perception; Fin. Prcep.: Fine-grained Perception; Ins. Reas.: Instance Reasoning; Logic. Reas.: Logical Reasoning; Math.: Mathematics; Sci. \& Tech.: Science \& Technology.}
\vspace{-3mm}
\label{tab:variant_r_mmstar}
\begin{adjustbox}{max width=\columnwidth}
\begin{tabular}{@{}lccccccc@{}}
\toprule
% Variant & Coa. Prcep. & Fin. Prcep. & Ins. Reas. & Logic. Reas. & Math. & Sci. \& Tech. & Avg. \\
% \multirow{2}{*}{Variant} & \multicolumn{7}{c}{MMStar} \\
% \cmidrule(lr){2-8}
Variant & Coa. Prcep. & Fin. Prcep. & Ins. Reas. & Logic. Reas. & Math. & Sci. \& Tech. & Avg. \\
\midrule
$r=0.1$          & 65.18 & 42.40 & 61.18 & 53.72 & 47.97 & 30.92 & 50.23 \\
$r=0.2$          & 67.16 & 42.70 & 63.85 & 51.66 & 49.23 & 32.30 & 51.15 \\
$r=0.3$          & 66.76 & 44.05 & 63.13 & 50.53 & 34.44 & 34.39 & 48.88 \\
$r=0.5$          & 65.61 & 43.44 & 63.68 & 48.91 & 37.66 & 27.10 & 47.73 \\
$r=1$            & 69.78 & 47.81 & 61.46 & 44.86 & 33.77 & 33.21 & 48.48 \\
only $\mathcal{S}^t$ & 66.69 & 43.44 & 63.82 & 52.60 & 50.22 & 36.87 & 52.27 \\
only $\mathcal{S}^v$ & 66.26 & 42.68 & 63.07 & 51.55 & 48.24 & 37.49 & 51.54 \\
\rowcolor{gray!15}
MGSI            & \best{71.17} & 43.89 & \best{67.40} & \best{57.31} & \best{52.11} & 36.73 & \best{54.77} \\
\bottomrule
\end{tabular}
\end{adjustbox}
\vspace{-3mm}
\end{table}

% \begin{wraptable}{r}{0.55\linewidth}
\begin{table}[t]
% \vspace{-13pt}
\centering
\footnotesize
\setlength{\tabcolsep}{4pt}
% \vspace{-2mm}
\renewcommand{\arraystretch}{1.08}
% \vspace{-3mm}
\caption{Comparison of different variants of SAB-LVLM on Qwen2.5-VL-32B-Instrust at MMStar benchmark. Coa. Prcep.: Coarse Perception; Fin. Prcep.: Fine-grained Perception; Ins. Reas.: Instance Reasoning; Logic. Reas.: Logical Reasoning; Math.: Mathematics; Sci. \& Tech.: Science \& Technology.}
\vspace{-3mm}
\label{tab:variant_sab}
\resizebox{\columnwidth}{!}{%
\begin{tabular}{@{}lccccccc@{}}
\toprule
% \multirow{2}{*}{Variant} & \multicolumn{7}{c}{MMStar} \\
% \cmidrule(lr){2-8}
Variant & Coa. Prcep. & Fin. Prcep. & Ins. Reas. & Logic. Reas. & Math. & Sci. \& Tech. & Avg. \\
\midrule
w/o SAB            & 69.78 & 47.81 & 61.46 & 44.86 & 33.77 & 33.21 & 48.48 \\
w/ SAB-R & 65.79 & 42.52 & 59.27 & 51.05 & 46.16 & 30.29 & 49.18 \\
w/ SAB-C & 66.97 & 45.83 & 63.80 & 54.39 & 48.88 & 34.47 & 52.39 \\
\rowcolor{gray!15}
\textbf{w/ SAB}            & \best{71.17} & 43.89 & \best{67.40} & \best{57.31} & \best{52.11} & \best{36.73} & \best{54.77} \\
\bottomrule
\end{tabular}
}
\vspace{-5mm}
\end{table}

\begin{figure*}
    \centering
    \includegraphics[width=\textwidth]{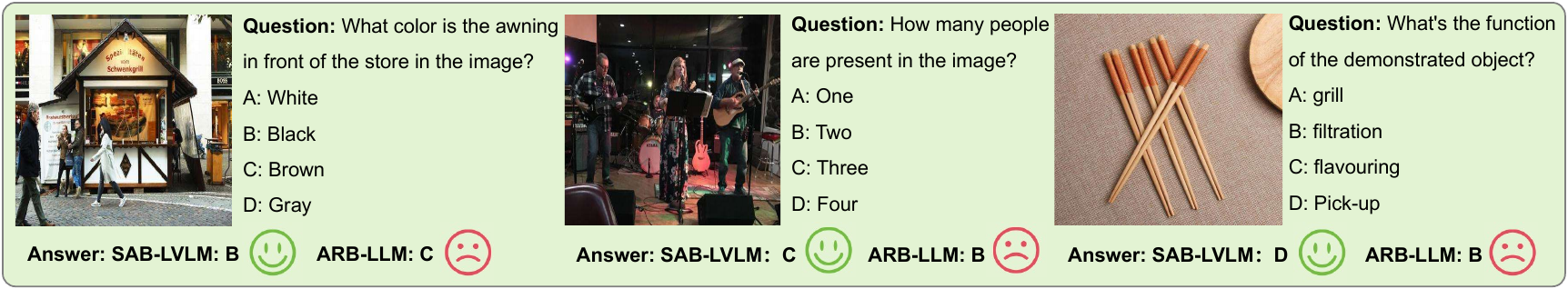}
    \vspace{-5mm}
    \caption{Qualitative analysis at MMStar with Qwen2.5-VL-7B-Instruct.}
    \label{mmstar_vis}
    \vspace{-3mm}
\end{figure*}

In Tab.~\ref{tab:qwen25_quant_results}, SAB-LVLM consistently outperforms all 1-bit baselines across all five benchmarks while maintaining an approximate 1-bit quantization depth.
Notably, PB-LLM~\cite{yuan2024pbllm} and BiLLM~\cite{huang2024billm} suffer severe performance degradation after binarization, indicating that naive partial binarization or reliance solely on residuals is insufficient for LVLM. In contrast, the proposed SAB-LVLM achieves stable gains across perception, document understanding, and video understanding tasks.
In Tab.~\ref{tab:internvl35_8b_main_results}, we present results across the InternVL3.5 family. The proposed SAB-LVLM significantly outperforms existing 1-bit baselines in most benchmarks, indicating our approach is independent of specific backbone networks.
In Tabs.~\ref{tab:qwen25_mmstar_results} and ~\ref{tab:qinternvl35_mmstar_results}, we report the detailed results on the MMStar benchmark across Qwen2.5-VL and InternVL3.5 families to further examine which capabilities are preserved. Compared to the 1-bit baselines, the proposed SAB-LVLM better preserves coarse perception and reasoning capabilities. 
In Tab~\ref{tab:vsibench_detailed_qwen}, we present the detailed results of VSI-Bench on Qwen2.5-VL-32B-Instruct. SAB-LVLM outperforms baseline methods across multiple tasks and achieves comparable performance to full-precision models in object appearance order, object absolute distance, and route planning tasks.
In Fig.~\ref{mmstar_vis}, We present qualitative analysis on MMStar using QwenVL-32B-VL. The three questions in the figure respectively tested the perception, counting, and reasoning capabilities of LVLM. Compared to ARB-LLM~\cite{chen2025hbllm}, SAB-LVLM successfully answered all three questions.

\subsection{Ablation Study}

In this section, we analyze the proposed components, including the spatial significance map, modality-guided significance integration (MGSI), and different variants of SAB-LVLM. All experiments were conducted on the MMStar benchmark using Qwen2.5-VL-32B, and detailed performance metrics are reported.

\subsubsection{Analysis of Spatial Significance Map.}

As shown in Tab.~\ref{tab:ablation_tau_mmstar}, we evaluate the performance of SAB-LVLM on Qwen2.5-VL-32B using three different values of the threshold $\tau$ on the MMStar benchmark. The results show that the spatial significance map is sensitive to the choice of $\tau$. Among all settings, $\tau=0.0003$ achieves the best overall performance, reaching 54.77 and improving the baseline by 6.29 points. In particular, this setting yields more evident gains on instance reasoning, logical reasoning, mathematics, and science \& technology, indicating that an appropriate threshold helps preserve multimodal reasoning ability under extreme low-bit compression. When $\tau$ is too small or too large, the distinction between weights activated under a single modal and those activated activated across modalities becomes less effective, which limits the benefit of significance-aware binarization map. These results indicate that partitioning spatial significance map using an appropriate threshold $\tau$ is crucial for preserving the capabilities of LVLMs.

\subsubsection{Analysis of Modality-Guided Significance Integration.}

\begin{figure}[h]
% \vspace{-3mm}
    \centering
    \begin{minipage}[t]{0.49\linewidth}
        \centering
        \includegraphics[width=\textwidth]{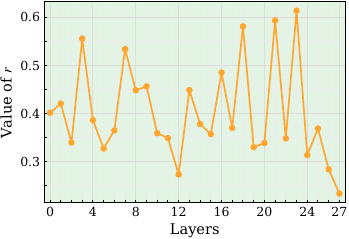}
        % \caption{$r$ of different self-attention output projection layers.}
    \end{minipage}\hfill
    \begin{minipage}[t]{0.49\linewidth}
        \centering
        \includegraphics[width=\textwidth]{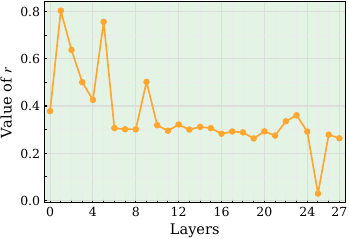}
        % \caption{$r$ of different down projection layers.}
    \end{minipage}
    \vspace{-3mm}
    \caption{Visualization of modality integration score $r$ among self-attention output projection layers (\emph{the top left}), the down projection layers (\emph{the top right}).
    }
    \vspace{-5mm}
\label{fig:r_layer}
\end{figure}

As shown in Tab.~\ref{tab:variant_r_mmstar}, we evaluate the performance of MMStar on Qwen2.5-VL-32B-Instrust with different variants of modality-guided significance integration. Using a fixed modality integration score $r$ yields limited improvements. The best fixed setting ($r=0.2$) reaches 51.15, still below the MGSI (54.77). 
As shown in Fig.~\ref{fig:r_layer}, we present the distribution of $r$ among different layers, including self-attention output projection, down projection layers.
Since $r_\ell$ varies significantly across different layers, simply fixing $r$ yields limited effectiveness.
Therefore, adaptively estimating $r$ for each layer is more effective than using a global constant.
Moreover, using only a single-modality sensitivity map (only $\mathcal{S}^t$ or only $\mathcal{S}^v$) still underperforms the joint MGSI (52.27/51.54 vs. 54.77), confirming that LVLM binarization requires jointly modeling both modalities rather than treating them independently.  

% \vspace{-3mm}
\subsubsection{Analysis of different variants of SAB-LVLM.}
As shown in Tab.~\ref{tab:variant_sab}, we report MMStar results on Qwen2.5-VL-32B-Instruct under different variants of SAB-LVLM. ``w/o SAB'' denotes the baseline, which adopts the same alternating refinement strategy as~\cite{li2025arbllm}. ``w/ SAB-R'' applies \emph{row-wise} significance reweighting in the alternating significance-weighted update, where all weights within the same row share an identical significance value, i.e., significance value of $i$-th row $\boldsymbol{\Lambda}_{ij}
=\frac{1}{n}\sum_{j=1}^nr \cdot\mathbf{\Gamma}_{ij}+(1-r)\cdot(1-\mathbf{\Gamma}_{ij})$. ``w/ SAB-C'' applies \emph{column-wise} significance reweighting, where all weights within the same column share an identical significance value, i.e., significance value of $j$-th column $\boldsymbol{\Lambda}_{ij}
=\frac{1}{m}\sum_{i=1}^m r \cdot\mathbf{\Gamma}_{ij}+(1-r)\cdot(1-\mathbf{\Gamma}_{ij})$. Finally, ``w/ SAB'' uses the proposed significance map defined in Eq.~\eqref{eq:lambda}. As shown in Tab.~\ref{tab:variant_sab}, the complete spatial significance map (w/ SAB) achieved the highest average score.

\section{Conclusion}

In this paper, we introduce \textbf{SAB-LVLM}, a novel significance-aware binarization framework for large vision-language models, which serves as an early exploration of weight binarization for LVLMs. Specifically, we first construct a spatial significance map using multimodal calibration data to distinguish full-precision weights activated under a single modality from those activated across modalities. We then devise a modality-guided significance integration strategy to measure weight significance across layers and modalities, and further build a significance-aware binarization map. Based on this design, we employ an alternating significance-weighted update scheme to perform LVLM binarization under approximate 1-bit constraints. Extensive experiments on representative LVLM benchmarks demonstrate that SAB-LVLM consistently outperforms existing binarization methods across multiple downstream tasks. In the future work, we will evaluate the proposed SAB-LVLM on a broader set of LVLM architectures, more diverse multimodal tasks, or longer-context video understanding scenarios.

\bibliographystyle{IEEEtran}
\bibliography{main}
\end{document}